\documentclass[journal]{IEEEtran}
\ifCLASSINFOpdf
\else
\fi
\usepackage{amsmath,amsfonts}
\usepackage{algorithmic}
\usepackage{algorithm}
\usepackage{array}
\usepackage[caption=false,font=normalsize,labelfont=sf,textfont=sf]{subfig}
\usepackage{textcomp}
\usepackage{stfloats}
\usepackage{url}
\usepackage[hidelinks]{hyperref}
\usepackage{xurl}
\usepackage{verbatim}
\usepackage{graphicx}
\usepackage{cite}
\usepackage{multirow}
\usepackage{placeins}

\begin{document}
\setlength{\textfloatsep}{8pt}
\setlength{\intextsep}{8pt}
%
\title{Temporal Image Representations for Mouse Dynamics Authentication}
%
%
%

\author{Junyi Li,
        Charles Devlen,
        and~Daqing Hou,~\IEEEmembership{Senior Member,~IEEE}%
\thanks{J. Li is with the Department of Computer Science and Engineering,
Washington University in St. Louis, St. Louis, MO 63130.
C. Devlen is with the PhD Department of Computing and Information Sciences.
D. Hou is with the Department of Software Engineering.
They are located at the Rochester Institute of Technology,
Rochester, NY 14623.}%
}

%
%

\markboth{}{}
%



\maketitle

\begin{abstract}
Mouse dynamics authentication has traditionally relied on handcrafted features or time-series representations that capture both spatial and temporal characteristics of mouse movements. More recent approaches have modeled mouse data as images, enabling the use of powerful pretrained computer vision models. However, prior work primarily focuses on spatial-only representations where the cursor coordinates are plotted as a trajectory image. These representations discard the temporal characteristics which can help discriminate users. In this work, we propose image representations that incorporate speed and velocity information for mouse dynamics authentication. We investigate two mouse data representations, a XYPlot representation that intuitively shows the path of movement, and a Pairwise Distance (PWD) representation that utilizes the entire image canvas to encode relationships between mouse events. Both representations are evaluated under both location-aware and location-free settings using pretrained 2D-CNN and ViT architectures. We found that the incorporation of temporal information improved performance over spatial representations alone, and that the location-aware variants generally outperformed the location-free. We achieve state-of-the-art performance on two datasets, Balabit and TWOS, and extend our evaluation to two other datasets. We found that in general, location-aware PWD with temporal encoding combined with the 2D-CNN architecture provided the best results, with mean EERs of 5.21\% and 12.70\% on the Balabit Test-1 and Test-2 sets, respectively, 4.94\% on TWOS, 6.16\% on ChaoShen, and 0.59\% on DFL.
\end{abstract}


%
\IEEEpeerreviewmaketitle

\section{Introduction}
The protection of personal data becomes increasingly important as digital systems become more entwined with our daily lives. Recent large-scale data breaches \cite{the_united_states_department_of_justice_2017, the_united_states_department_of_justice_2020, moab2024} have allowed malicious actors to compromise personal data through the use of traditional username/password login credentials. These breaches highlight the need for additional factors of authentication to protect personal data. However, commonly used additional factors, such as one-time passcodes and push requests for approval, introduce high user friction due to frequent requests and the burden of enrolling in these services. Additionally, most of these factors only challenge the user a single time at login, rather than continuously monitoring devices in the background.

Behavioral biometrics address the need for unobtrusive and continuous authentication by profiling the unique characteristics of users as they interact with their devices. Mouse dynamics is a category of behavioral biometrics that seeks to distinguish users by their unique mouse usage patterns. Many prior works rely on handcrafted features combined with traditional machine learning methods in order to classify users \cite{shen2012continuous, 10.1145/2046707.2046725, gamboa2003, 10.1145/1966913.1966983, antal2019, askari2009, shen2012effectiveness, shen2014anomaly}. More recent works have explored deep learning approaches for mouse dynamics, allowing models to automatically extract relevant features from the input, reducing the need for manual feature extraction and increasing performance \cite{chong2018mouse, chong2020, hu2019insider, fu2020rumba, AntalDeepLearning, 9465583}. 

Prior deep learning approaches for mouse dynamics modeling can be divided into two main categories: time-series modeling approaches \cite{chong2020, AntalDeepLearning, 9465583, fu2020rumba}, and computer vision approaches \cite{chong2018mouse, chong2020, hu2019insider}. Motivated by the success of a pretrained 2D-CNN in prior work \cite{chong2018mouse, chong2020}, we focus on computer vision methods for mouse dynamics authentication. These methods focus on modeling mouse data as images, allowing established vision architectures to be used for authentication. However, converting mouse event sequences into images introduces important design considerations that determine which characteristics of mouse movement are preserved. 

Computer vision mouse dynamics approaches often render mouse trajectories as 2D coordinate plots, which we refer to as XYPlot representations. Although these images preserve the spatial path of a trajectory, they do not explicitly encode how the cursor moves over time. Consequently, inputs with similar spatial paths but different movement speeds or velocities may appear similar to the model.
\IEEEpubidadjcol
Temporal information has been incorporated into XYPlot images for related tasks, such as enhancing mouse bot detection \cite{niu2023exploring, wei2019deep}, but its contribution has not been systematically examined alongside other representation choices. In addition, while some works incorporate information about where the cursor moved on screen \cite{chong2018mouse, chong2020}, others choose to center trajectory images, discarding potentially useful information about where user interactions occurred on screen \cite{niu2023exploring, wei2019deep}. The effects of temporal information and absolute cursor position, both individually and in combination with image representation, therefore warrant further investigation.

In this work, we systematically evaluate three factors affecting image-based mouse dynamics authentication: image representation, temporal information, and cursor location information. We compare conventional sparse XYPlot representations with a dense Pairwise Distance (PWD) image representation, which encodes pairwise distances between mouse events across the image canvas. For each representation, we investigate the addition of two temporal features, speed and velocity, and compare location-aware variants which retain absolute cursor position with location-free variants. We further consider both location-aware variants, which retain absolute cursor position, and location-free variants to determine the effect of including screen occupancy information.

We evaluate our proposed image representations using pretrained 2D Convolutional Neural Networks (2D-CNN) and Vision Transformer (ViT) based authentication frameworks across four publicly available mouse dynamics datasets. Since performance based on individual samples was found to be impractical for authentication, we employ a score-level fusion strategy that aggregates predictions from multiple input samples. Performance is reported using Equal Error Rate (EER) and Area Under the Curve (AUC) as a function of the number of mouse events needed to make a decision. 

Our results demonstrate that all three of these factors affect authentication performance. PWD generally outperforms XYPlot, and incorporating speed or velocity generally improves performance over spatial-only representations. Location-aware representations also generally outperform their corresponding location-free variants. We further observe interactions between these factors, including differences in the effectiveness of speed and velocity encoding between XYPlot and PWD representations.

The contributions of this work are as follows:
\begin{enumerate}
    \item We define a set of image representation methods that examine the effect of using XYPlot vs. PWD, temporal information, and location information.

    \item We achieve state-of-the-art performance among vision-based mouse dynamics authentication approaches across four public datasets, extending the two datasets in \cite{chong2020}.

    \item We demonstrate that the trends in results for our proposed image representations generalize well across both 2D-CNN and Vision Transformer architectures.
\end{enumerate}


This paper is organized as follows: Section \ref{section:Related Work} examines prior methods for mouse dynamics authentication. Section \ref{section:Temporal Encoding Framework} describes the methodology used in our experiment for segmentation and evaluation. Section \ref{section:Architectures} describes the architecture of our model. Section \ref{section:Experimental Setup} defines our experimental setup and protocol. Section \ref{section:Results} details the results obtained from our experimentation. Section \ref{section:Discussion} discusses our results and the relationships we observed between image representations and model performance. Finally, Section \ref{section:Conclusion} concludes our paper.


\section{Related Work}
\label{section:Related Work}

\begin{table*}[t]
\caption{Representation methods, corresponding papers, and method descriptions.}
\centering
\begin{tabular}{|p{0.15\textwidth}|p{0.15\textwidth}|p{0.2\textwidth}|p{0.4\textwidth}|}
\hline
\textbf{Method} & \textbf{Description} & \textbf{Papers} & \textbf{Paper Details} \\
\hline

\multirow{2}{0.15\textwidth}{Handcrafted Feature Vectors} &
\raggedright\arraybackslash Features are manually calculated from each sample and represented as a vector
& Shen et al. \cite{shen2012continuous} & \raggedright\arraybackslash Utilized 92 total features derived from click timing, movement speed, and acceleration. \\
\cline{3-4}
& & Shen et al. \cite{shen2012effectiveness} & \raggedright\arraybackslash 13 total features based on distance, time, speed, and acceleration. \\
\cline{3-4}
& & Aksari and Artuner \cite{askari2009} & \raggedright\arraybackslash 144 total features based on speed, deviation, angle, and acceleration. \\
\cline{3-4}
& & Gamboa and Fred \cite{gamboa2003} & \raggedright\arraybackslash 63 features from x-y coordinates, path distance, angle, curvature, velocity, tangential velocity, tangential acceleration, tangential jerk, and angular velocity. \\
\cline{3-4}
& & Ahmed and Traore \cite{ahmed2007new} & \raggedright\arraybackslash 39 total features derived from speed, direction, action type, traveled distance, and elapsed time. \\
\cline{3-4}
& & Khan and Hou \cite{khan2023user} & \raggedright\arraybackslash 32 features from traveled distance, curve length, duration, speed, acceleration, and angle. \\
\hline

\multirow{2}{0.15\textwidth}{Time-Series Modeling} &
\raggedright\arraybackslash Models sequential mouse events as sequences of timesteps
& Chong et al. \cite{chong2020} & \raggedright\arraybackslash Two time-series feature sets were evaluated in this work, ($\Delta x$, $\Delta y$), and ($\frac{\Delta x}{\Delta t}$, $\frac{\Delta y}{\Delta t}$). Both fixed-length sequences of 130 events and variable-length sequences were evaluated. \\
\cline{3-4}
& & Fu et al. \cite{fu2020rumba} & \raggedright\arraybackslash Time series sequences with features ($\Delta x$, $\Delta y$, $v_x$ and $v_y$) were utilized for authentication. \\
\hline

\multirow{2}{0.15\textwidth}{Image Encoding } &
\raggedright\arraybackslash Transforms sequential mouse events into images, often leveraging spatial properties of the data
& Chong et al. \cite{chong2020} & \raggedright\arraybackslash A 448$\times$448 image was generated with x-y coordinates plotted as a line. \\
\cline{3-4}
& &  Wei et al. \cite{wei2019deep} & \raggedright\arraybackslash Plotted each trajectory as an unconnected sequence of points on a 600$\times$600 image with the Red and Green channels used for distance magnitude and time delta. \\
\cline{3-4}
& &  Hu et al. \cite{hu2019insider} & \raggedright\arraybackslash Plotted each trajectory as a line, coloring each sample based on action type (drag-and-drop, point-and-click) and assigning point markers to click events and periods of inactivity. \\
\cline{3-4}
& &  Niu et al. \cite{niu2023exploring} & \raggedright\arraybackslash Encoded features such as velocity, acceleration, time and angle as color and point size onto trajectory plots for use in bot detection. \\
\hline

\end{tabular}
\label{table:representation_categories}
\end{table*}
In this section we review mouse dynamics approaches, as summarized by Table \ref{table:representation_categories}, and discuss their relevance to mouse dynamics image-based authentication.

Early mouse dynamics authentication approaches primarily relied on handcrafted feature vectors extracted from raw mouse event data. These handcrafted features were designed to describe different attributes of user interaction behavior, including distance, velocity, acceleration, angle, and duration \cite{shen2012continuous, shen2012effectiveness, askari2009, gamboa2003, ahmed2007new, khan2023user, khan2024}. Such approaches have demonstrated that handcrafted features can be effective for mouse dynamics authentication. However, their effectiveness depends on domain expertise in the formulation and selection of the features, which may limit their ability to capture complex relationships present within each user's behavior.

More recent works have introduced deep learning approaches that learn representations directly from lower-level input features derived from the raw mouse events. These methods generally use either time-series representations or image-based representations. In time-series approaches, input samples are represented as a sequence of timesteps, with each timestep consisting of coordinate differences or horizontal and vertical velocity components. Chong et al. \cite{chong2020} and Fu et al. \cite{fu2020rumba} evaluated sequences based on coordinate differences $(\Delta x, \Delta y)$ and velocity components $(\frac{\Delta x}{\Delta t}, \frac{\Delta y}{\Delta t})$. These approaches allow sequential models to learn temporal dependencies directly from the input data and leverage information about how the movement evolved over time.

An alternative approach represents mouse trajectories as images and applies computer vision models for authentication. This allows mouse dynamics authentication to be treated as a computer vision problem, making it possible to leverage large-scale pretrained models developed for computer vision. For example, Chong et al. \cite{chong2020} renders mouse coordinates as XY plots relative to screen resolution, allowing a pretrained multilabel 2D-CNN to learn behavioral representations from spatial patterns. Their work demonstrated that a pretrained 2D-CNN can learn discriminative patterns from mouse images without the need for manually engineered features. Other works such as Hu et al. \cite{hu2019insider} rendered mouse actions as images, using color channels and point marker styles to signal different user interaction behaviors.

A key design choice in computer vision mouse dynamics is which information is encoded in the image. We consider three related factors: temporal information, image representation, and location information.

First, existing image representations differ in the extent to which they encode temporal information. Conventional XYPlot representations encode \textit{where} the cursor moved, rather than \textit{how} it moved over time. Since the input images are rendered according to cursor coordinates, timing characteristics such as inter-event time differences, velocity, and acceleration are not encoded within the image. Two trajectories that have similar spatial paths may have very different timing and movement speed, which is not captured by this representation. This limitation is highlighted by the use of temporal features in both traditional handcrafted features approaches as well as time-series representations.  Other mouse trajectory image representations have incorporated additional temporal information to improve performance. Wei et al. \cite{wei2019deep} represent mouse samples as sequences of points and encode distance and timing information through color channels, while Niu et al. \cite{niu2023exploring} encoded kinematic information such as velocity, acceleration, time, and angle into the color channels and through the radius of trajectory points. However, both of these works focus on bot detection rather than user authentication, and the individual contribution of features is not explored, focusing only on the best performing combinations of features.

A second limitation is the utilization of the image canvas by current representation methods. XYPlot images only encode information along the rendered trajectory, while most of the image consists of uninformative background pixels. This approach requires vision models to extract discriminative information from a small portion of the image, potentially limiting performance. Chong et al. \cite{chong2018mouse} examined the information used by 2D-CNNs when processing XYPlot images, and found that they use curve information sparsely, with a tendency to assign little relevance to straight segments and line crossings. These findings motivate the exploration of denser image representations that utilize the full image canvas, and preserve information that may otherwise be lost when multiple mouse events occur at the same coordinates.

Finally, whether or not the absolute cursor position is preserved in the final image representation varies across prior works. Image-based mouse representations can either incorporate the location of the cursor on screen \cite{chong2018mouse, chong2020}, or discard it \cite{niu2023exploring, wei2019deep}. Location-aware representations include where the user moved on screen by plotting relative to the screen resolution, potentially revealing user-specific patterns, such as a tendency to move the cursor in the upper-left region of the screen. Location-free representations focus only on the shape and overall structure of movement instead of where it occurred, centering trajectories on plots. The inclusion of where the user tends to operate on screen may affect model performance, motivating further exploration.

In this work, we build upon these prior approaches by systematically investigating the effects of image representation, temporal encoding, and location information on mouse dynamics authentication. Rather than replacing the spatial cursor coordinate representation, our approach retains the spatial structure of mouse movement while encoding speed and velocity information into the image representation. Additionally, we propose a novel pairwise distance representation which does not depend on the sparse cursor coordinate representation, utilizing the full image canvas and without creating artifacts when a mouse event occurs at the same coordinates as another in a given sample. We evaluate this pairwise representation both alone and combined with speed and velocity information, and present location-aware and location-free variants of our representations.

\section{Methodology}
\label{section:Temporal Encoding Framework}
In this section we describe the transformation of the raw data to image representations. The formulation for both XYPlot and Pairwise Distance images is described, as well as the method used to encode speed and velocity for each.

\subsection{Raw Mouse Data and Preprocessing}
Raw mouse data consists of three main components: cursor coordinates, timestamps, and event type. Cursor coordinates describe the position of the mouse cursor on screen. Each coordinate is associated with a timestamp indicating when the cursor occupied that position, as well as an event type such as ``move", ``press", or ``release". The ``press" and ``release" event type correspond to mouse button interactions. Let a mouse event sequence be represented as $\mathcal{S}=\{e_1,\ldots,e_M\}$, where $e_i=(x_i,y_i,t_i)$ denotes the $i$-th event, with cursor coordinates $(x_i,y_i)$ and timestamp $t_i$.

Before segmentation, data cleaning was performed for each dataset. Duplicate records in the data were dropped by timestamp, retaining only the first, and only the ``move" events were retained to focus on active movement. Records with invalid coordinates ($x$ or $y \geq 65535$) were removed, and non-numeric values in the coordinate and timestamp fields were discarded.

The cleaned data were segmented into non-overlapping windows of 125 events, matching the average sequence length reported by the best-performing segmentation strategy in prior work \cite{chong2020}, while providing a fixed input size. The fixed sequence length is particularly important for the proposed pairwise distance representation, which produces an $M\times M$ matrix. All resulting representations are converted to a resolution of $448\times448$ for use as input samples.

\subsection{Baseline Image Representations}
We construct two baseline image representations from each mouse sequence: XYPlot and Pairwise Distance (PWD). XYPlot renders consecutive cursor coordinates as connected line segments, whereas PWD encodes the pairwise Euclidean distances in an $M \times M$ matrix.

For the location-free XYPlot, each trajectory input is normalized using its local minimum and maximum cursor coordinates per segment in the range ([0,1]). The longer side is then scaled to our final image size of 448 pixels while preserving the aspect ratio and the trajectory is centered along the shorter dimension. For the location-aware XYPlot, the x and y coordinates are normalized to ([0,1]) using a common, user-specific scaling factor derived from the user's estimated screen resolution, preserving the screen's estimated aspect ratio. The rendered trajectory occupies a rectangular region that reflects the estimated screen resolution. Symmetric white padding is then applied along the shorter dimension to obtain a square image representation.

The pairwise distance representation is constructed as a $M \times M$ distance matrix $D$, where each element stores the Euclidean distance between events $i$ and $j$ is $D_{ij} = \sqrt{(x_i-x_j)^2+(y_i-y_j)^2}$, where $i,j = 1,...,M$. The resulting distance matrix can be interpreted as an image representation of the mouse event sequence. The rows and columns of the matrix correspond to the order of events within the sequence, and each element of the matrix $D_{ij}$ stores the Euclidean distance between events $i$ and $j$. The pairwise distance matrix can be expressed as:
\begin{equation}
D=
\begin{bmatrix}
D_{11} & D_{12} & \cdots & D_{1M} \\
D_{21} & D_{22} & \cdots & D_{2M} \\
\vdots & \vdots & \ddots & \vdots \\
D_{M1} & D_{M2} & \cdots & D_{MM}
\end{bmatrix}
\end{equation} 
The pairwise distance matrix satisfies $D_{ij}=D_{ji}$ since the Euclidean distance is symmetric. The diagonal elements correspond to the distance of each event to itself, so $D_{i=j}=0$.

For the location-free version of PWD, the distances within each fixed-length sample are normalized by the maximum pairwise distance within that sample. This removes the influence of the absolute cursor position and scale of the movement while preserving the relative spatial relationships between events. The normalized PWD matrix is then mapped to grayscale intensity and resized to the final image size of 448$\times$448.

For the location-aware PWD representation, the pairwise distances are instead normalized using the estimated diagonal of the user's screen. The red and blue channels are used to store the normalized distance matrix, while the green stores normalized x and y coordinates. 

\begin{figure}[!t]
\centering

\includegraphics[width=0.40\columnwidth]{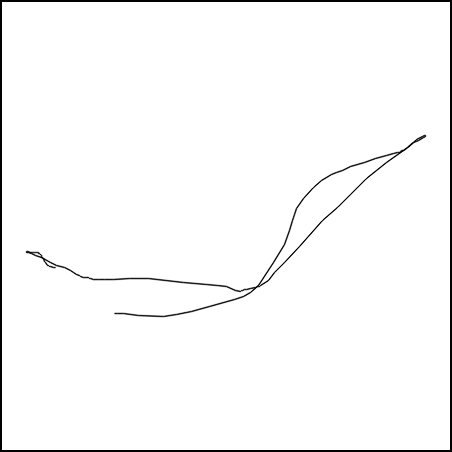}
\hspace{0.02\columnwidth}
\includegraphics[width=0.40\columnwidth]{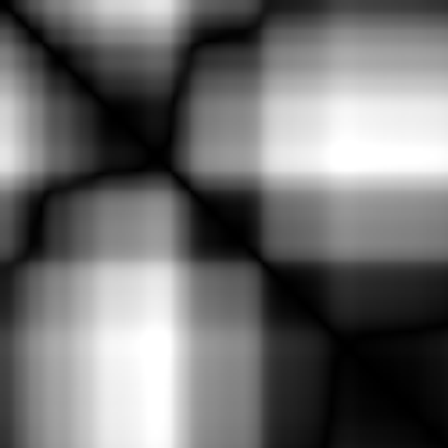}

\vspace{1mm}

\makebox[0.40\columnwidth][c]{%
    \shortstack{(a) Location-Free\\XYPlot}}
\hspace{0.06\columnwidth}
\makebox[0.40\columnwidth][c]{%
    \shortstack{(b) Location-Free\\PWD}}

\vspace{2mm}

\includegraphics[width=0.40\columnwidth]{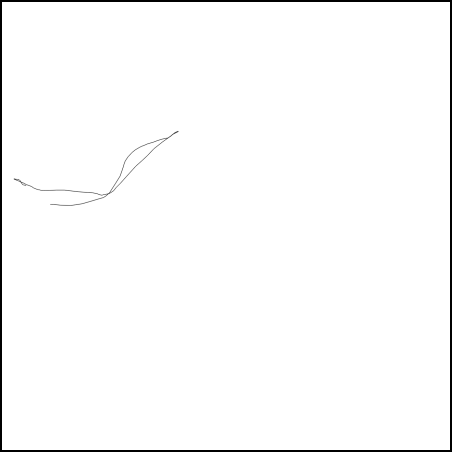}
\hspace{0.02\columnwidth}
\includegraphics[width=0.40\columnwidth]{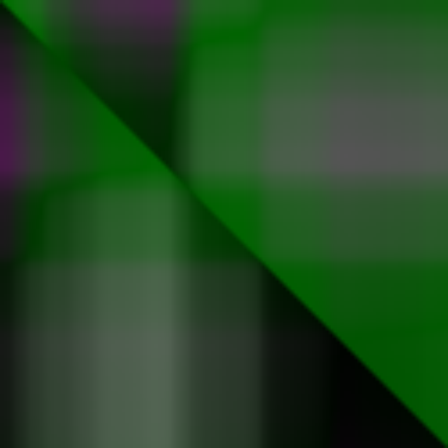}

\vspace{1mm}

\makebox[0.40\columnwidth][c]{%
    \shortstack{(c) Location-Aware\\XYPlot}}
\hspace{0.06\columnwidth}
\makebox[0.40\columnwidth][c]{%
    \shortstack{(d) Location-Aware\\PWD}}

\caption{
Comparison of location-free and location-aware image representations generated
from the same mouse event sequence in the Balabit dataset. The images correspond
to the first 125-event chunk extracted from Session 0041905381 of User 7 using
the fixed-length chunking strategy.
}
\label{fig:xyplot_pairwise}

\end{figure}

The green channel is divided into two triangular regions, where the upper and lower regions encode the normalized x- and y-coordinates, respectively. This normalization preserves the relationship between the observed mouse movements and the estimated spatial scale of the user's screen. The resulting matrix is mapped to the image representation and resized to $448\times448$ pixels. Therefore, while the location-free PWD representation removes absolute spatial scale by normalizing within each sequence, the location-aware representation retains information about movement scale relative to the user's estimated screen dimensions. Figure~\ref{fig:xyplot_pairwise} illustrates the XYPlot and the PWD generated from the same input sample. 

\subsection{Movement Attributes}
We derive three movement attributes derived from the raw mouse event sequence: speed, horizontal velocity, and vertical velocity. Let two consecutive mouse events be given by $e_i=(x_i,y_i,t_i)$ and $e_{i+1}=(x_{i+1},y_{i+1},t_{i+1})$ with change of components as: $\Delta x_i = x_{i+1}-x_i$, $\Delta y_i = y_{i+1}-y_i$, and $\Delta t_i = t_{i+1}-t_i$. The definitions of the three movement attributes are summarized in Table~\ref{tab:temporal_features}.

\begin{table}[!t]
\caption{Movement Attributes Used for Velocity Encoding}
\label{tab:temporal_features}
\centering
\begin{tabular}{l l}
\hline
Feature ($\Phi$) & Definition \\
\hline
Speed ($v$)
&
$\sqrt{\Delta x_i^2+\Delta y_i^2}/\Delta t_i$
\\

Horizontal Velocity ($v_x$)
&
$\Delta x_i/\Delta t_i$
\\

Vertical Velocity ($v_y$)
&
$\Delta y_i/\Delta t_i$
\\
\hline
\end{tabular}
\end{table} 
Speed represents cursor speed magnitude independent of direction. In contrast, horizontal and vertical velocities reflect cursor motion along the respective coordinate axes and preserve directional information through their signs. Together, these two attributes are referred to as \textit{velocity}. Speed and velocity are incorporated into the image representations through the encoding framework introduced in the following subsection.

\subsection{Velocity Representation}
\label{sec:velocity_representation}

We extend the baseline XYPlot and PWD representations by encoding speed or velocity into their RGB channels. Each attribute is normalized to ([0,1]) using the minimum and maximum values computed from the training set, then multiplied by 255 to produce pixel intensities ([0,255]). Given a velocity attribute $\phi$, the corresponding velocity attribute sequence is represented as $\Phi=[\phi_1,\phi_2,\ldots,\phi_M]$.

\begin{figure}[h]
\centering
\label{fig:xyplot_pairwise_vxvy}

\includegraphics[width=0.40\columnwidth]{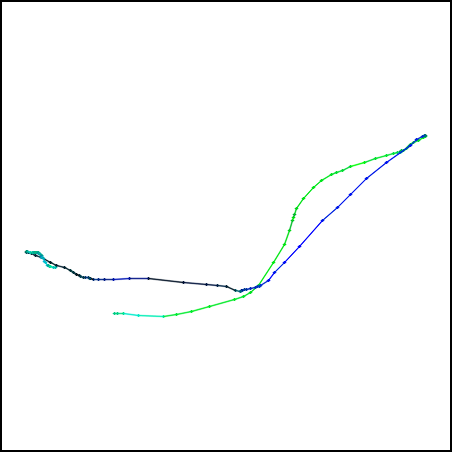}
\hspace{0.02\columnwidth}
\includegraphics[width=0.40\columnwidth]{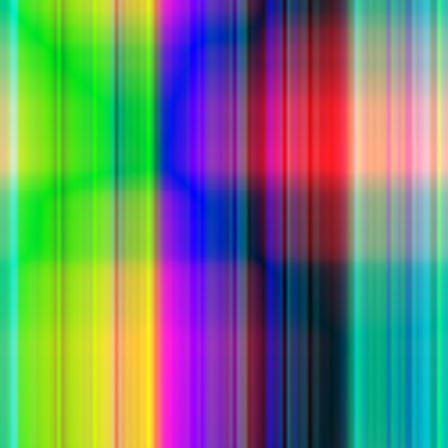}

\vspace{1mm}

\makebox[0.40\columnwidth][c]{%
    \shortstack{(a) Location-Free\\Velocity-Encoded XYPlot}}
\hspace{0.06\columnwidth}
\makebox[0.40\columnwidth][c]{%
    \shortstack{(b) Location-Free\\Velocity-Encoded PWD}}

\vspace{2mm}

\includegraphics[width=0.40\columnwidth]{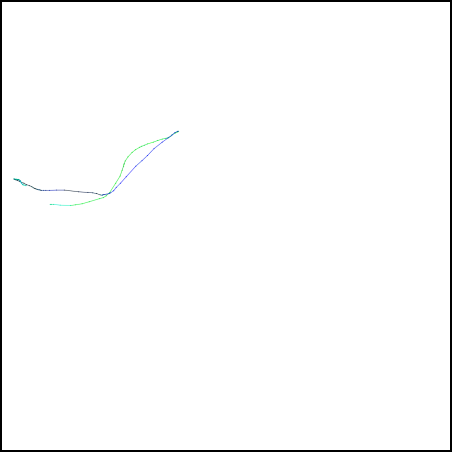}
\hspace{0.02\columnwidth}
\includegraphics[width=0.40\columnwidth]{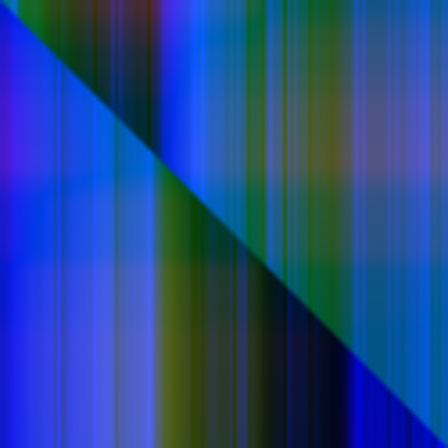}

\vspace{1mm}

\makebox[0.40\columnwidth][c]{%
    \shortstack{(c) Location-Aware\\Velocity-Encoded XYPlot}}
\hspace{0.06\columnwidth}
\makebox[0.40\columnwidth][c]{%
    \shortstack{(d) Location-Aware\\Velocity-Encoded PWD}}

\caption{
Comparison of location-free and location-aware velocity-encoded image
representations generated from the same mouse event sequence in the Balabit
dataset. The images correspond to the first 125-event chunk extracted from
Session 0041905381 of User 7 using the fixed-length chunking strategy.
}
\label{fig:velocity_encoded_representations}

\end{figure}

\noindent\textit{a) Cursor Coordinate Representation:} The rendered trajectory can be represented as a binary trajectory mask:

\begin{equation}
M(x,y)=
\begin{cases}
1, & (x,y)\in\mathcal{T}\\
0, & \text{otherwise}
\end{cases}
\end{equation} where $\mathcal{T}$ denotes the set of pixels occupied by the rendered cursor trajectory. To associate trajectory pixels with mouse events, we define a mapping $\pi:\mathcal{T}\rightarrow\{1,2,\ldots,M\}$, where $\pi(x,y)$ returns the index of the mouse event associated with the rendered trajectory pixel $(x,y)$. The velocity-encoded RGB image is then defined as:

\begin{equation}
I(x,y)=
\begin{cases}
\left(
0,
\phi_{\pi(x,y)}^{(1)},
\phi_{\pi(x,y)}^{(2)}
\right),
&
M(x,y)=1\\
(1,1,1),
&
M(x,y)=0
\end{cases}
\end{equation}
Thus, velocity information is encoded only along the rendered trajectory. Figure\nobreak\hspace{0.3em}\ref{fig:velocity_encoded_representations}(a) and Figure\nobreak\hspace{0.3em}\ref{fig:velocity_encoded_representations}(c) illustrates the velocity-encoded cursor coordinate representation (XYPlot), where the red channel is fixed, while the horizontal and vertical velocity components ($v_x$ and $v_y$) are assigned to the green and blue channels, respectively.

\noindent\textit{b) Pairwise Distance Representation:} For the Pairwise Distance Representation, the velocity attribute sequence is transformed into an event-indexed feature matrix. The column index corresponds to the mouse event index, and each velocity attribute is replicated across all rows to construct the velocity feature channel: $T_{ij}=\phi_i$ with $i,j=1,\ldots,M$, where $T$ denotes the velocity attribute matrix. The matrix can be expressed as:
\begin{equation}
T=
\begin{bmatrix}
\phi_1 & \phi_2 & \cdots & \phi_M \\
\phi_1 & \phi_2 & \cdots & \phi_M \\
\vdots & \vdots & \ddots & \vdots \\
\phi_1 & \phi_2 & \cdots & \phi_M
\end{bmatrix}
\end{equation} 

The final velocity-encoded Pairwise Distance Representation is constructed by combining the pairwise distance matrix with two velocity attribute matrices. Let $D$ denote the Pairwise Distance Matrix Representation introduced in the previous section, and let $T^{(1)}$ and $T^{(2)}$ denote the event-indexed feature matrices corresponding to the two normalized velocity components. The RGB representation is defined as $\tilde{I}_{ij} = \left(D_{ij}, T_{ij}^{(1)}, T_{ij}^{(2)} \right)$ with $i,j=1,\ldots,M$. This $M\times M$ three-channel representation is then resized to a fixed spatial resolution of $448\times448$, producing the final image $I(x,y)$ used as the network input. Figure~\ref{fig:velocity_encoded_representations}(b) illustrates the velocity-encoded pairwise distance representation.

\begin{figure*}[t]
\centering
\includegraphics[width=2.0\columnwidth]{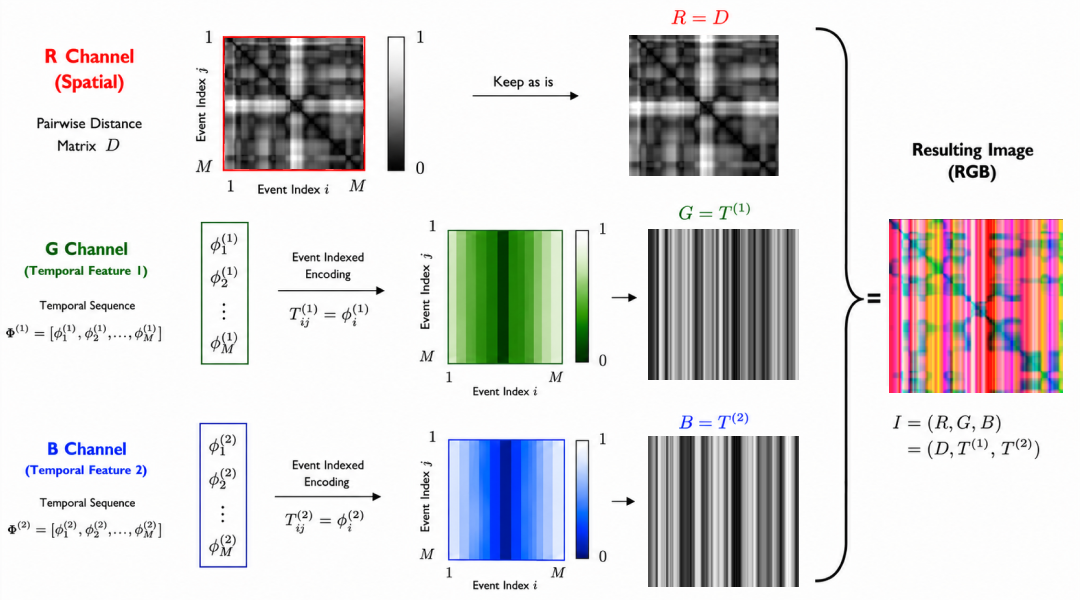}
\caption{
Example of the proposed velocity encoding framework. The pairwise distance matrix is assigned to the red channel, while horizontal velocity ($v_x$) and vertical velocity ($v_y$) are encoded as temporal stripe images and assigned to the green and blue channels, respectively. The three channels are subsequently combined to form the final temporal-encoded image representation. The illustrated sample corresponds to a 125-events chunk extracted from Session 1909471574 of User 35 in the Balabit dataset.
}
\label{fig:temporal_framework}
\end{figure*}

Figure~\ref{fig:temporal_framework} illustrates the proposed velocity-encoded Pairwise Distance Representation. The pairwise distance matrix is assigned to the red channel. The two event-indexed feature matrices are assigned to the green and blue channels, respectively. These three channels are then combined to produce the final RGB image representation.

\subsection{Image Representation Schemes}
\label{sec:image_representation_schemes}
\begin{table}[!t]
\caption{Image Representation Schemes and Channel Assignments}
\label{tab:encoding_schemes}
\centering
\footnotesize

\begin{tabular}{llccc}
\hline
Setting & Schemes & Red & Green & Blue \\
\hline

\multirow{6}{*}{Location-Free}
& (1) & $M$ & $M$ & $M$ \\
& (2) & $M$ & $M_{|v|}$ & $M_{|v|}$ \\
& (3) & $M$ & $M_{v_x}$ & $M_{v_y}$ \\
& (4) & $D$ & $D$ & $D$ \\
& (5) & $D$ & $T_v$ & $T_v$ \\
& (6) & $D$ & $T_{v_x}$ & $T_{v_y}$ \\
\hline

\multirow{6}{*}{Location-Aware}
& (7) & $\hat{M}$ & $\hat{M}$ & $\hat{M}$ \\
& (8) & $\hat{M}$ & $\hat{M}_{|v|}$ & $\hat{M}_{|v|}$ \\
& (9) & $\hat{M}$ & $\hat{M}_{v_x}$ & $\hat{M}_{v_y}$ \\
& (10) & $\hat{D}$ & $T_{\hat{x},\hat{y}}$ & $\hat{D}$ \\
& (11) & $\hat{D}$ & $T_{\hat{x},\hat{y}}$ & $T_v$ \\
& (12) & $\hat{D}$ & $T_{\hat{x},\hat{y}}$ & $T_{v_x,v_y}$ \\
\hline
\noalign{\vskip 4pt}

\end{tabular}
\vspace{4pt}
\parbox{0.96\columnwidth}{%
\footnotesize
\textit{Note:} $M$ and $\hat{M}$ denote the location-free and location-aware
trajectory masks; $D$ and $\hat{D}$ denote the corresponding pairwise
distance matrices. Subscripts indicate the feature information encoded
in the corresponding mask or matrix: $|v|$, $v_x$, and $v_y$ denote speed,
horizontal velocity, and vertical velocity, respectively, while $\hat{x}$
and $\hat{y}$ denote the screen coordinates normalized according to the
estimated screen coordinate range. $T$ denotes the corresponding
event-indexed feature matrix.
}

\end{table}

We investigate twelve image representation schemes in total, summarized in Table \ref{tab:encoding_schemes}. Schemes (1)--(6) correspond to the location-free setting, while schemes (7)--(12) correspond to the location-aware setting. Within each setting, the first three schemes are constructed from the cursor coordinate representation, while the remaining three are constructed from the pairwise distance representation. For each representation, the three schemes correspond to the original representation, speed encoding, and velocity encoding, respectively. Each scheme is used as model input in our evaluation.

\section{Model Architectures}
\label{section:Architectures}
To model behavioral patterns from mouse trajectory data, we employ two widely used visual backbone architectures: a convolutional neural network (2D-CNN) and a Vision Transformer (ViT). The models treat the authentication task as a visual pattern recognition problem. They take input representations as 448×448 RGB images, from which the models learn discriminative behavioral patterns.

For user authentication, we adopt a multi-label classification formulation. This design is partially inspired by the authentication framework proposed by Chong et al.~\cite{chong2020}. Instead of training a separate model for each user, we employ a shared feature extraction backbone followed by multiple independent output nodes. Each node acts as a user-specific binary classifier that evaluates whether the input sample belongs to that user, producing an independent score in the range $[0,1]$.

\subsection{Multi-CNN Architecture}
\begin{figure}[t]
\centering
\includegraphics[
    width=0.5\textwidth,
    height=0.15\textheight
]{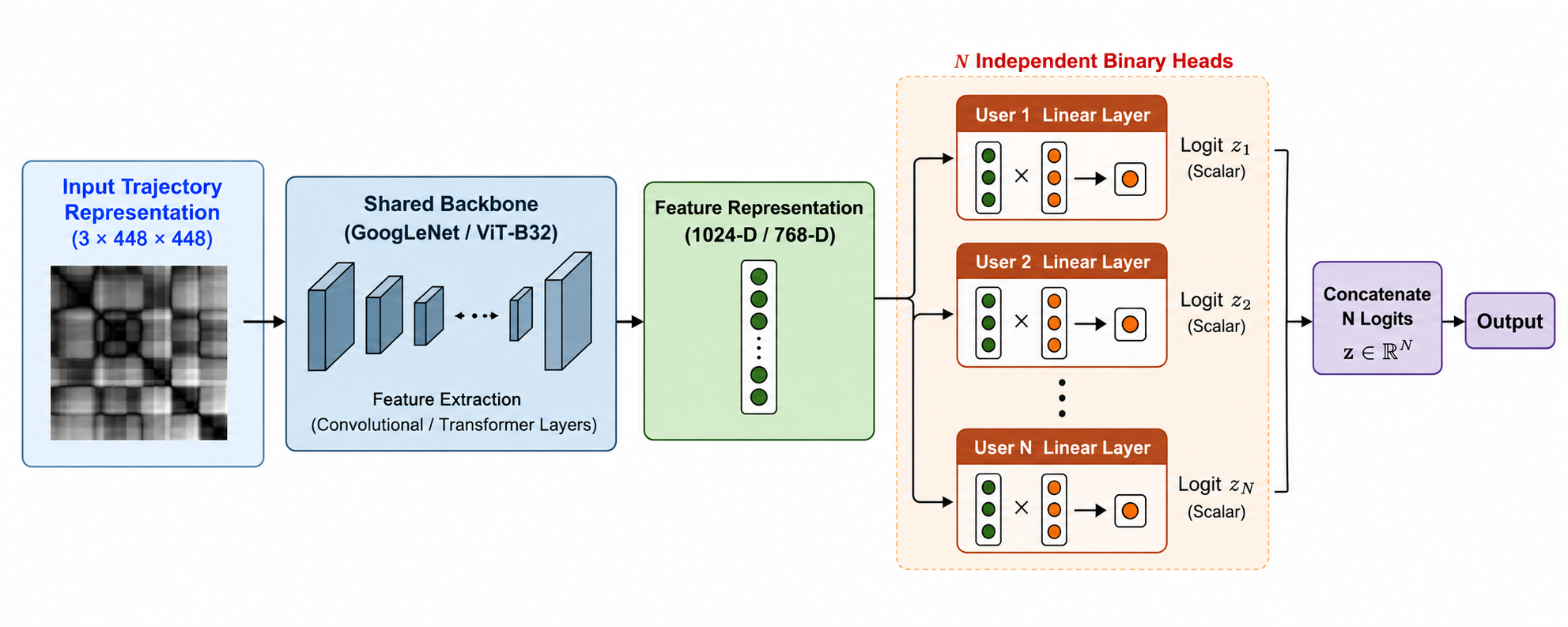}
\caption{
Multi-label authentication framework.
A shared backbone (GoogLeNet or ViT-B/32)
extracts image features, which are passed to user-specific binary classification heads.
}
\label{fig:model_arch}
\end{figure}

Although mouse dynamics data naturally exhibits temporal structure, the relatively small size of existing datasets makes training deep time-series models from random initialization challenging. Following Chong et al.~\cite{chong2020}, we adopt a shared multi-label 2D-CNN architecture based on a GoogLeNet backbone initialized with ImageNet pretrained weights~\cite{GoogleNet} from PyTorch package \cite{paszke2019pytorch} as an instance of 2D-CNN model. A single shared backbone is used to extract behavioral representations across all users, while user-specific classification heads are attached for authentication. Instead of training $n$ independent binary 2D-CNN models for $n$ users, this architecture maintains all user-specific heads within a single shared network, significantly reducing computational overhead and the number of trainable parameters through shared feature extraction layers. Each user-specific head consists of an independent fully connected layer that maps the shared 1024-dimensional feature representation to a binary authentication score. Unlike conventional multi-class classification, our proposed architecture performs independent binary authentication for each user, where the outputs of different heads do not compete with each other for classification. The outputs of all user-specific heads are concatenated to form the final score vector. Each head independently produces a score in the range $[0,1]$ indicating the likelihood that the input sample belongs to the corresponding user. During inference, authentication is performed using only the head associated with the claimed user. The overall architecture of the proposed Multi-CNN model is illustrated in Fig.~\ref{fig:model_arch}. 

\subsection{Multi-ViT Architecture}

To evaluate whether the proposed temporal image representations generalize beyond convolutional neural networks, we additionally employ a Vision Transformer (ViT) backbone. Unlike 2D-CNNs, which learn local spatial patterns through convolutional operations, ViTs model global relationships between image regions using self-attention mechanisms \cite{Dosovitskiy2020AnIIViT}. This provides a fundamentally different visual learning paradigm and allows us to examine whether the benefits of temporal image encoding are architecture-independent.

Specifically, we adopt a ViT-B/32 architecture initialized with ImageNet-1K pretrained weights. The backbone partitions each input image into non-overlapping $32 \times 32$ patches, which are subsequently projected into a learnable embedding space and processed by a stack of Transformer encoder layers. Following the standard ViT formulation, a learnable classification token is prepended to the patch sequence and serves as the global image representation after Transformer encoding \cite{Dosovitskiy2020AnIIViT}. All image representations are resized to $448 \times 448$ pixels prior to model input. Although the pretrained backbone was originally trained using $224 \times 224$ images, the patch size remains fixed at $32 \times 32$, resulting in a $14 \times 14$ patch grid during fine-tuning. Position embeddings are automatically interpolated to accommodate the increased input resolution. The pretrained weights are provided through the timm library \cite{rwightman2019timm}. The original ImageNet classification head is removed and replaced with the same multi-label authentication architecture described in Section IV-A. The shared ViT backbone produces a 768-dimensional feature representation for each input image, which is passed to user-specific binary classification heads. Similar to the Multi-CNN architecture, each head independently predicts the likelihood that the input sample belongs to the corresponding user, and the outputs of all heads are concatenated to form the final authentication score vector.

The overall framework differs from the Multi-CNN model only in the choice of feature extraction backbone. This design enables a direct evaluation of whether the proposed temporal image representations provide consistent benefits across both convolutional and transformer-based architectures.

\subsection{Training Objective}

Given an input sample, the model produces a vector of logits 
$\mathbf{z} \in \mathbb{R}^{U}$, where $U$ denotes the number of users. 
Each element $z_u$ corresponds to the authentication score for user $u$. 
During inference, a sigmoid function is applied to convert logits into 
independent scores in the range $[0,1]$. Training labels are represented using one-hot encoding, where $\mathbf{y} \in \{0,1\}^{U}$ indicates the user identity of the ground-truth.

For loss function, we employ a weighted binary cross-entropy loss applied independently to each user head. Let $p_u = \sigma(z_u)$ denote the predicted probability for user $u$. The loss for each user head is defined as \[
\ell_u =
- \left(
w_1 y_u \log p_u +
w_2 (1-y_u)\log(1-p_u)
\right),
\]where $w_1$ and $w_2$ are weighting factors for positive and negative 
samples respectively. The overall training objective is obtained by 
averaging the losses across all user heads:
\[
\mathcal{L} =
\frac{1}{U}\sum_{u=1}^{U}\ell_u .
\]The weights are computed using the empirical positive and negative 
ratios in the training set to mitigate the severe class imbalance 
inherent in multi-user authentication tasks. Specifically,

\[
w_1 = \frac{1}{{F_1}}, \qquad
w_2 = \frac{1}{F_2},
\] where $F_1$ and $F_2$ denote the empirical positive and negative sample 
ratios in the training set. The models were implemented using PyTorch 1.21 and trained on an NVIDIA A100 GPU. Training was performed using the AdamW optimizer with a learning rate of 0.0001. A batch size of 20 was used for the 2D-CNN models, while a batch size of 256 was used for the ViT models. All models were trained for 25 epochs, with the learning rate decayed by a factor of 0.1 every 7 epochs. Model checkpoints were selected based on the best validation average EER, with early stopping applied using a patience value of 5.

\section{Experimental Setup}
\label{section:Experimental Setup}
\subsection{Datasets}
Our experiments use four publicly available datasets Balabit \cite{balabit_2016}, TWOS \cite{harilal2017twos}, ChaoShen \cite{shen2012effectiveness} and DFL \cite{antal2019}.

The Balabit mouse dynamics challenge \cite{balabit_2016} was released for an online mouse dynamics authentication competition. The dataset consists of sessions from 10 users, collected as they completed their everyday work tasks remotely. Mouse data is stored in separate training and testing directories, and contains information such as x and y coordinates, the type of event, button pressed, and timestamp. The testing portion of the Balabit dataset contains both ``legal" (genuine) and ``illegal" (impostor) sessions. 

The TWOS dataset \cite{harilal2017twos} was collected in a constrained setting where 24 people participated in a gamified stock trading competition designed for data collection. Each session consists of x-y coordinates, the type of event, timestamp, and screen resolution information.

The ChaoShen dataset \cite{shen2012continuous} was collected in laboratory conditions from 28 users completing their everyday tasks. Participant activities consisted of Internet surfing, word processing, online chatting, programming, and online games. Each participant used the same mouse and monitor during collection. Each session of data collection lasted approximately 30 minutes. Each session contains x and y coordinates, timestamps, the type of event, and the ID of the process users were in.

The DFL dataset \cite{antal2019} was collected from 21 users on their own devices, with 20 laptops and 1 desktop. Like ChaoShen, the participants were unconstrained in their activities. Some users used both a trackpad and an external mouse during data collection. Each session consists of x-y coordinates, the button pressed, the type of event, and the timestamp.

These datasets represent a wide range of operating condition, from task-constrained (Balabit, TWOS) to completely unconstrained (ChaoShen, DFL). This allows for a realistic measure of performance to be determined in our evaluations.

\subsection{Training and Testing Procedure}
In our evaluation we adopt the training and testing procedure from \cite{chong2020, antal2019}. Our evaluation is sample based, meaning that authentication decisions by the model are rendered based on a single input sample. We found that single sample performance was impractical for authentication, and accordingly we adopted a score-level fusion technique that averaged scores for multiple samples.

For the Balabit dataset, learning was conducted exclusively on the ``training" portion of the dataset, and two test protocols were defined, Test-1 and Test-2. The Test-1 protocol uses all the legal sessions of the users for testing, where the target users sessions are labeled as genuine, and all other users as impostor. Test-2 uses both legal and illegal sessions, labeling the target user’s legal sessions as genuine, and the illegal sessions for the target user as impostor. Due to the smaller total quantity of data, we used oversampling of the positive class for balancing. 

For the TWOS, ChaoShen, and DFL datasets, the training set consists of the first 2/3 of data in each session, while the testing dataset consists of last 1/3 of data for each session.

\section{Results}
\label{section:Results}

In this section, we evaluate the authentication performance of the proposed image representations. Performance is reported using Equal Error Rate (EER) and Area Under the Curve (AUC). We evaluate both single-sample authentication and aggregated authentication using score-level fusion, where scores from multiple samples are fused to produce a single authentication decision for each user. To determine whether differences in EER were statistically significant between experiments, we conducted a Repeated-Measures Analysis Of Variance (RM-ANOVA) test, followed by pairwise post-hoc comparisons with Holm correction at a significance level of $\alpha=0.05$. 

\subsection{Single-Sample Authentication Performance}
We first evaluate the discriminative performance of single input samples. Tables \ref{tab:single_sample_MultiCNN} and \ref{tab:single_sample_MultiViT} report the mean EER and AUC for the encoding schemes outlined in Table \ref{tab:encoding_schemes} using the Multi-CNN and Multi-ViT architectures, respectively. We organize the results around the three factors we identified from our literature survey in Section \ref{section:Related Work}: image representation (XYPlot or PWD), temporal encoding (speed or velocity) and location information (location-aware or location free). 

\begin{table*}[t]
\caption{Single-Sample Authentication Performance of Different Image Representations\\ 
Using Multi-CNN with 125 Events per Sample}
\label{tab:single_sample_MultiCNN}
\centering
\scriptsize
\begin{tabular*}{\textwidth}{@{\extracolsep{\fill}}lcccccccccc}

\hline

\multirow{4}{*}{\textbf{Representation}} 
& \multicolumn{10}{c}{\textbf{Datasets}}\\

\cline{2-11}

& \multicolumn{4}{c}{\textbf{Balabit}} 
& \multicolumn{2}{c}{\textbf{TWOS}} 
& \multicolumn{2}{c}{\textbf{ChaoShen}} 
& \multicolumn{2}{c}{\textbf{DFL}}\\

\cline{2-5}
\cline{6-7}
\cline{8-9}
\cline{10-11}
\noalign{\vskip 2pt}

& \multicolumn{2}{c}{\textbf{EER (\%)}} 
& \multicolumn{2}{c}{\textbf{AUC}}
& \multirow{2}{*}{\textbf{EER (\%)}} 
& \multirow{2}{*}{\textbf{AUC}}
& \multirow{2}{*}{\textbf{EER (\%)}} 
& \multirow{2}{*}{\textbf{AUC}}
& \multirow{2}{*}{\textbf{EER (\%)}} 
& \multirow{2}{*}{\textbf{AUC}}\\

\cline{2-5}

& \textbf{Test-1} & \textbf{Test-2}
& \textbf{Test-1} & \textbf{Test-2}
& & & & & & \\

\hline

\multicolumn{11}{@{}l}{\textbf{Location-Free (Centering)}}\\
\hline

(1) XYPlot & 14.84 & 20.67 & 92.51 & 85.64 & 20.51 & 86.90 & 17.76 & 90.45 & 9.85 & 96.34\\
(2) XYPlot + Speed & 9.63 & 16.11 & 96.33 & 90.47 & 11.97 & 94.10 & 12.48 & 94.64 & 3.92 & 99.20\\
(3) XYPlot + Velocity & 10.99 & 16.76 & 95.24 & 89.43 & 13.60 & 92.89 & 12.84 & 94.39 & 4.42 & 99.04\\
(4) PWD & 10.61 & 16.48 & 95.64 & 90.17 & 11.75 & 94.08 & 12.33 & 94.71 & 4.51 & 99.00\\
(5) PWD + Speed & 7.89 & 14.23 & 97.29 & 91.80 & 10.89 & 94.60 & 9.52 & 96.00 & \textbf{2.41} & \textbf{99.57}\\
(6) PWD + Velocity & \textbf{6.67} & \textbf{13.12} & \textbf{98.04} & \textbf{92.65} & \textbf{10.88} & \textbf{94.67} & \textbf{8.80} & \textbf{96.87} & 2.45 & 99.55\\

\hline

\multicolumn{11}{@{}l}{\textbf{Location-Aware (Without Centering)}}\\
\hline

(7) XYPlot (Chong) & 7.17 & 23.55 & 97.58 & 82.00 & 7.06 & 97.04 & 14.42 & 92.61 & 0.82 & 99.91\\
(8) XYPlot + Speed & 5.33 & 19.69 & 98.29 & 86.30 & 6.12 & 97.61 & 8.92 & 96.63 & 0.76 & 99.93\\
(9) XYPlot + Velocity & 6.10 & 20.57 & 98.00 & 85.18 & 6.64 & 97.31 & 9.95 & 95.98 & 0.81 & 99.91\\
(10) PWD & 7.52 & 13.41 & 97.50 & 92.71 & 7.59 & 97.10 & 10.99 & 95.61 & 1.15 & 99.89\\
(11) PWD + Speed & 5.70 & 13.02 & 98.50 & 92.74 & \textbf{4.27} & \textbf{98.63} & 6.79 & 97.77 & \textbf{0.59} & \textbf{99.93}\\
(12) PWD + Velocity & \textbf{5.21} & \textbf{12.70} & \textbf{98.79} & \textbf{93.00} & 4.94 & 98.35 & \textbf{6.16} & \textbf{98.18} & 0.84 & 99.87\\

\hline
\end{tabular*}

\vspace{2pt}
\begin{minipage}{\textwidth}
\footnotesize
\textit{Note:} The pairs of representations with no statistically significant differences for each dataset are listed below. Balabit: (2,4), (7,10), (7,11), (8,10), (8,11); TWOS: (2,4), (2,5), (7,8), (7,10); ChaoShen: (2,4); DFL: (2,4), (7,9), (7,10), (7,11), (8,10), (8,11).
\end{minipage}

\end{table*}

\begin{table*}[t]
\caption{Single-Sample Authentication Performance of Different Image Representations\\ 
Using Multi-ViT with 125 Events per Sample}
\label{tab:single_sample_MultiViT}
\centering
\scriptsize
\begin{tabular*}{\textwidth}{@{\extracolsep{\fill}}lcccccccccc}

\hline

\multirow{4}{*}{\textbf{Representation}}
& \multicolumn{10}{c}{\textbf{Datasets}}\\

\cline{2-11}

& \multicolumn{4}{c}{\textbf{Balabit}}
& \multicolumn{2}{c}{\textbf{TWOS}}
& \multicolumn{2}{c}{\textbf{ChaoShen}}
& \multicolumn{2}{c}{\textbf{DFL}}\\

\cline{2-5}
\cline{6-7}
\cline{8-9}
\cline{10-11}
\noalign{\vskip 2pt}

& \multicolumn{2}{c}{\textbf{EER (\%)}}
& \multicolumn{2}{c}{\textbf{AUC}}
& \multirow{2}{*}{\textbf{EER (\%)}}
& \multirow{2}{*}{\textbf{AUC}}
& \multirow{2}{*}{\textbf{EER (\%)}}
& \multirow{2}{*}{\textbf{AUC}}
& \multirow{2}{*}{\textbf{EER (\%)}}
& \multirow{2}{*}{\textbf{AUC}}\\

\cline{2-5}

& \textbf{Test-1}
& \textbf{Test-2}
& \textbf{Test-1}
& \textbf{Test-2}
& & & & & & \\

\hline

\multicolumn{11}{@{}l}{\textbf{Location-Free (Centering)}}\\
\hline

(1) XYPlot & 17.48 & 23.41 & 89.67 & 82.37 & 25.35 & 82.21 & 29.29 & 78.20 & 14.93 & 93.10\\
(2) XYPlot + Speed & 11.62 & 20.33 & 94.70 & 85.86 & 14.28 & 92.21 & 13.57 & 94.09 & 4.71 & 98.92\\
(3) XYPlot + Velocity & 13.51 & 21.99 & 93.42 & 83.50 & 17.66 & 89.40 & 16.56 & 91.68 & 5.16 & 98.76\\
(4) PWD & 12.49 & 18.00 & 94.24 & 87.87 & 14.54 & 92.10 & 13.72 & 93.67 & 5.45 & 98.48\\
(5) PWD + Speed & 10.73 & 15.74 & 95.51 & 89.90 & \textbf{12.89} & \textbf{93.14} & 10.41 & 95.64 & \textbf{3.89} & \textbf{98.87}\\
(6) PWD + Velocity & \textbf{8.54} & \textbf{15.27} & \textbf{96.95} & \textbf{90.55} & 13.55 & 92.70 & \textbf{10.32} & \textbf{95.98} & 3.92 & 98.82\\

\hline

\multicolumn{11}{@{}l}{\textbf{Location-Aware (Without Centering)}}\\
\hline

(7) XYPlot (Chong) & 10.55 & 24.46 & 95.76 & 80.97 & 9.22 & 95.74 & 17.91 & 89.70 & 1.45 & 99.80\\
(8) XYPlot + Speed & 6.72 & 21.13 & 97.60 & 84.30 & 7.73 & 96.60 & 12.21 & 94.40 & 1.37 & 99.80\\
(9) XYPlot + Velocity & 7.28 & 23.16 & 97.48 & 82.33 & 8.19 & 96.42 & 14.79 & 92.61 & 1.44 & 99.79\\
(10) PWD & 8.95 & 18.08 & 96.59 & 88.16 & 12.96 & 93.54 & 17.15 & 90.72 & 2.63 & 99.55\\
(11) PWD + Speed & 7.04 & 16.10 & 97.67 & 90.27 & \textbf{5.86} & \textbf{97.89} & 7.93 & 97.40 & \textbf{0.71} & \textbf{99.93}\\
(12) PWD + Velocity & \textbf{5.80} & \textbf{16.05} & \textbf{98.21} & \textbf{88.97} & 5.96 & 97.74 & \textbf{7.57} & \textbf{97.50} & 0.80 & 99.87\\

\hline

\end{tabular*}

\vspace{2pt}
\begin{minipage}{\textwidth}
\footnotesize
\textit{Note:} The pairs of representations with no statistically significant differences for each dataset are listed below. Balabit: (2,4), (4,10), (7,10), (8,10), (8,11); TWOS: (2,4), (2,5), (4,10); ChaoShen: (2,4), (2,8), (3,9), (7,10); DFL: (7,8), (7,11), (8,11).
\end{minipage}

\end{table*}

\subsubsection{Effect of Image Representation} Under the location-free setting, PWD achieves statistically significantly lower EER than XYPlot across all four datasets for both the 2D-CNN and ViT architectures. This trend becomes less consistent when absolute location information is retained. For 2D-CNN, location-aware XYPlot achieves lower EER than PWD on Balabit Test-1, TWOS, and DFL, while PWD performs better on ChaoShen. The ViT results exhibit a different pattern: XYPlot performs better on TWOS and DFL, while PWD achieves lower EER on Balabit Test-1/Test-2 and ChaoShen. Overall, the results suggest that PWD is the more effective location-free representation, while the relative performance of XYPlot and PWD under the location-aware setting depends on both the dataset and the model architecture. This pattern suggests that incorporating absolute location information may benefit XYPlot more strongly, narrowing its performance gap with PWD. Compared with PWD and XYPlot with speed encoding in both location settings, PWD + Speed achieves lower EER than XYPlot + Speed in all four datasets for 2D-CNN and ViT except for Balabit Test-1. This suggests that the advantage between XYPlot and PWD still inherits when additional speed information was added.

\subsubsection{Effect of Temporal Information} 
Encoding speed information generally improves performance compared to the baseline image representations. For the location-free representations, XYPlot + Speed significantly outperforms XYPlot across all datasets and both architectures. Similarly, PWD + Speed significantly outperforms PWD across all datasets and both architectures. For the location-aware representations, XYPlot + Speed exhibits a similar trend, performing significantly better except on TWOS for 2D-CNN and DFL for ViT. PWD + Speed significantly outperforms PWD across all datasets and architectures. These findings suggest that the inclusion of temporal information increases the discriminability of the input samples, thereby improving performance.

\subsubsection{Effect of Location Information}

The location-aware (LA) setting generally outperforms the location-free (LF) setting across all representations on Balabit Test-1, TWOS, ChaoShen, and DFL for both 2D-CNN and ViT. For 2D-CNN, all the corresponding LF-LA comparisons in TWOS and DFL show statistically significant differences, while for ViT, statistically significant differences are shown on DFL. Furthermore, the performance improvement from incorporating location information is numerically greater for XYPlot representations than for PWD representations across these datasets and both architectures. This suggests that XYPlot representations benefit more from the additional location information. However, an opposite trend is found on Balabit Test-2, where the LF setting numerically outperforms the LA setting across all representations for both architectures, although none of these differences are statistically significant. One possible explanation is that the LA setting captures not only users' mouse movement patterns but also their preferred cursor activity regions on the screen. While such location information can provide additional discriminative cues under our current evaluation settings, it may not generalize well to previously unseen impostors. Since Balabit Test-2 involves unseen impostors, their cursor activity regions may differ from those learned during training, potentially limiting the generalization of LA representations compared with LF representations. Overall, these results demonstrate that incorporating location information can improve authentication performance, but its generalizability requires further investigation.

\subsection{Score-Level Fusion Performance}

\begin{figure*}[t]
\centering

\includegraphics[width=0.24\textwidth]{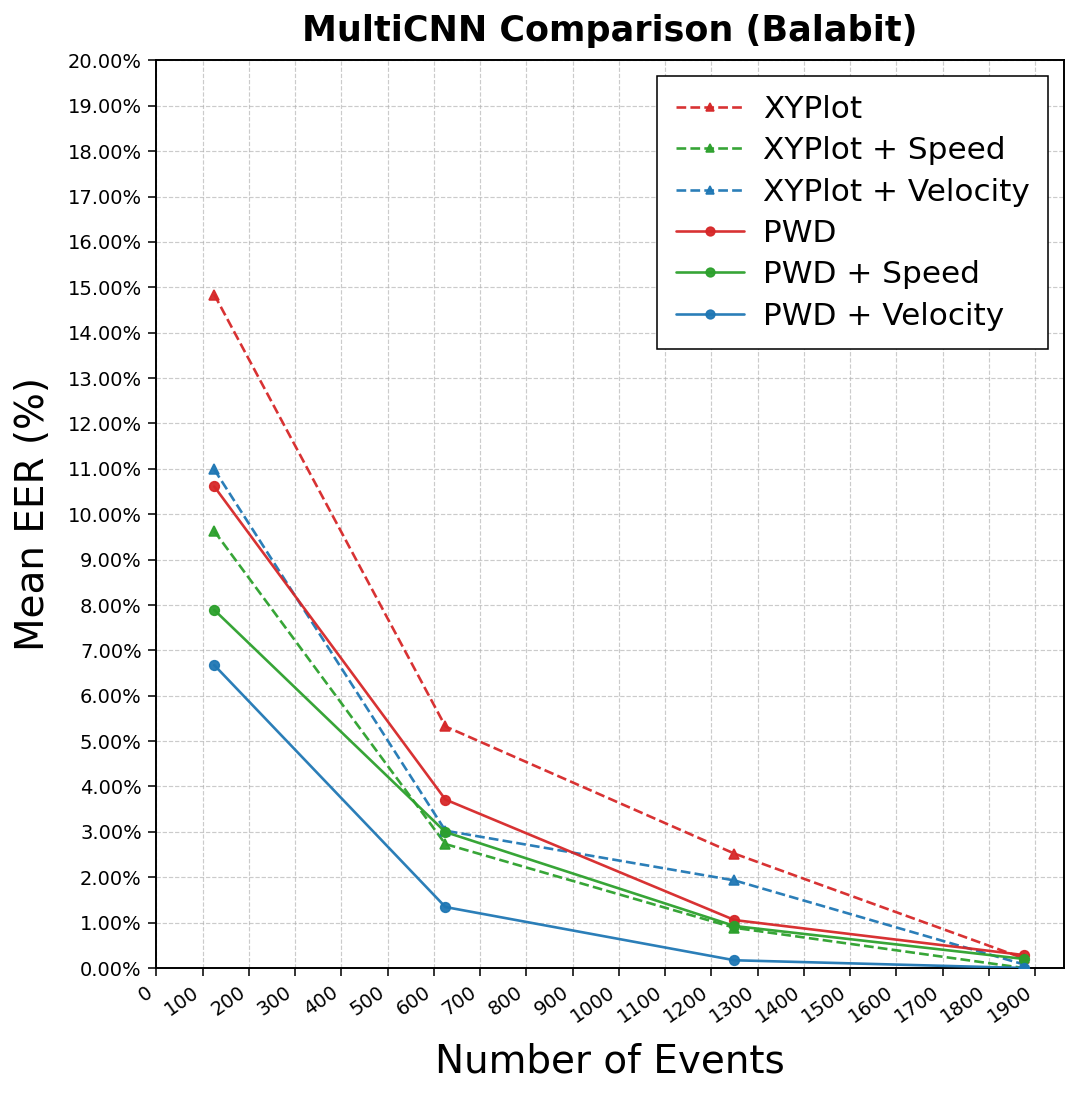}%
\hfill%
\includegraphics[width=0.24\textwidth]{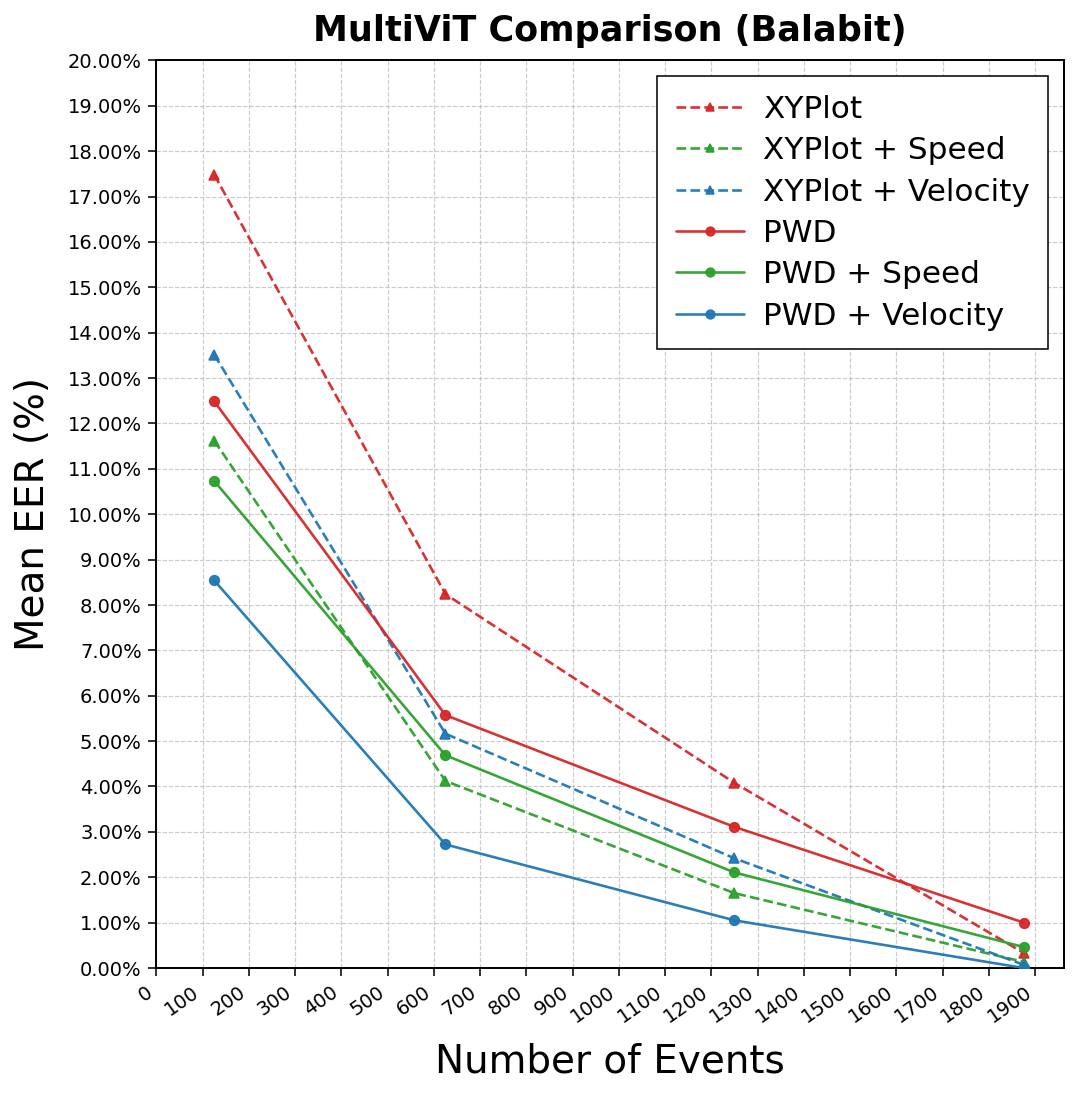}%
\hfill%
\includegraphics[width=0.24\textwidth]{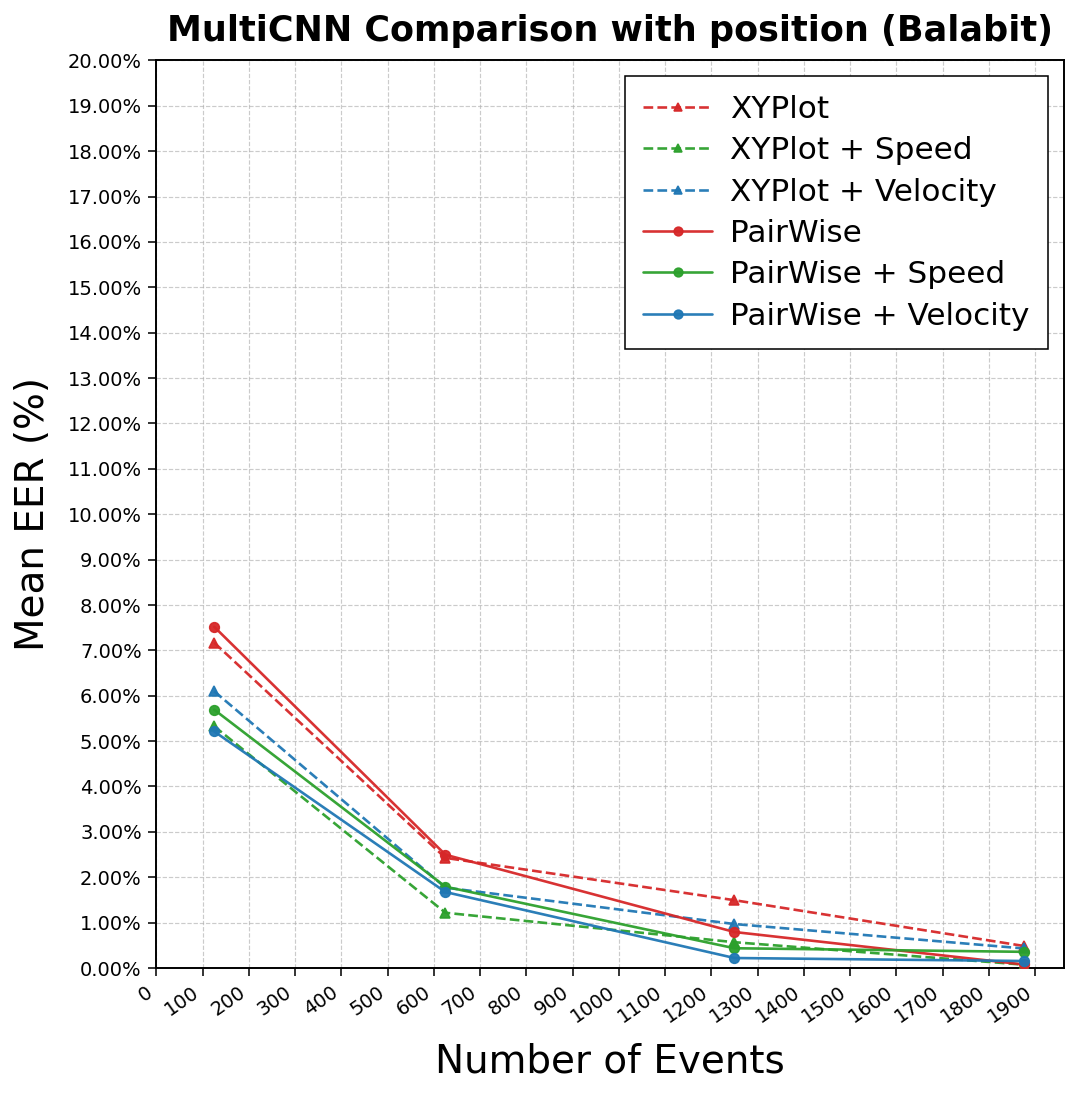}%
\hfill%
\includegraphics[width=0.24\textwidth]{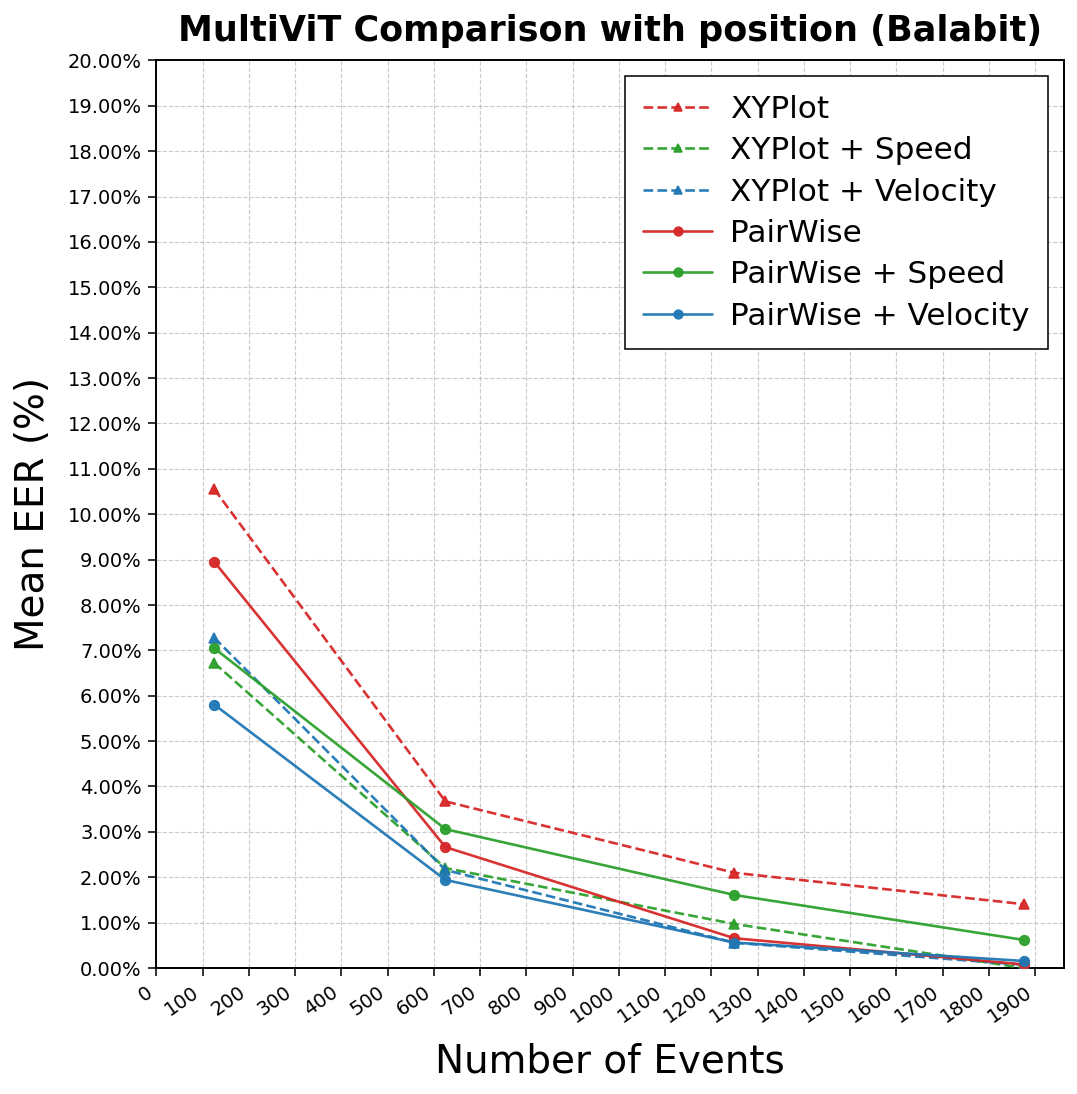}%

\vspace{1mm}

\makebox[0.24\textwidth][c]{\footnotesize (a) Multi-CNN: Location-Free}%
\hfill%
\makebox[0.24\textwidth][c]{\footnotesize (b) Multi-ViT: Location-Free}%
\hfill%
\makebox[0.24\textwidth][c]{\footnotesize (c) Multi-CNN: Location-Aware}%
\hfill%
\makebox[0.24\textwidth][c]{\footnotesize (d) Multi-ViT: Location-Aware}%

\caption{Score-level fusion performance on the Balabit dataset under location-free and location-aware schemes. Mean EER (\%) is reported as a function of the number of events used for score fusion. (a) Multi-CNN under the location-free scheme. (b) Multi-ViT under the location-free scheme. (c) Multi-CNN under the location-aware scheme. (d) Multi-ViT under the location-aware scheme.}
\label{fig:score_level_fusion}

\end{figure*}

While single-sample authentication evaluates individual mouse dynamics samples independently, aggregating information from multiple samples can provide more reliable authentication decisions in practical applications. We therefore apply score-level fusion techniques described in Section~\ref{section:Experimental Setup} to combine the prediction scores from multiple consecutive samples and evaluate authentication performance as more behavioral information is accumulated. This further examines whether the performance differences among image representations observed in the single-sample setting still persist after score-level fusion. Figure~\ref{fig:score_level_fusion} shows the mean EER as a function of the number of fused samples per authentication decision on the Balabit dataset. As similar trends were observed across all four datasets, Balabit is presented as a representative example.

Across all four datasets, score-level fusion consistently improves authentication performance over single-sample evaluation. As additional behavioral information is aggregated through score-level fusion, the performance gaps among different image representations gradually decrease. These differences are most evident when only a limited amount of behavioral information is available. In this setting, among representations without temporal encoding (XYPlot and PWD), PWD consistently outperforms XYPlot under both location schemes and with both model architectures. Beyond the choice of image representation family, incorporating either speed or velocity encoding further improves the performance of both XYPlot and PWD. The location scheme also has a clear impact on performance, with each image representation performing better in the location-aware scheme than its corresponding representation in the location-free scheme. Across these representations and location schemes, 2D-CNN generally achieves better overall performance than ViT.

\section{Discussion}
\label{section:Discussion}
The results of our experiments demonstrate that the addition of temporal information to mouse dynamics image representations improves authentication performance over spatial representations. For both XYPlot and PWD representations, encoding speed or velocity consistently resulted in performance gains over the spatial baselines across all four datasets. In addition, incorporating the position of the cursor on screen also improved performance compared to the location-free schemes, indicating that including where each trajectory occurred benefits distinguishability.

However, when comparing speed and velocity encoding for location-free XYPlot and PWD, the best performer for each was inconsistent. For XYPlot, speed encoding outperformed velocity encoding, while the opposite was true for PWD. One possible explanation is the interaction between the spatial distribution of encoded information and the structure of the temporal encoding. In XYPlot, velocity information is confined along the trajectory segments, so the horizontal and vertical velocity components are represented only along a small fraction of the overall image. Velocity encoding therefore introduces two distinct channel-specific patterns within an already sparse representation. In contrast, speed encoding encodes the same velocity magnitude in both color channels, producing a simpler and identical channel structure along the trajectory. This simpler representation may be easier to capture by the model when temporal information is spatially constrained.

For PWD, temporal information is distributed throughout the image canvas. This provides a larger area over which the model can exploit the complementary information encoded within the horizontal and vertical velocity components. In addition, PWD explicitly encodes distance, whereas for XYPlot the model has to extract it from the extent of the trajectory. DFL is the only exception, where PWD with speed encoding achieves a slightly lower EER than with velocity encoding,although the difference is not statistically significant. Because DFL users interact across two monitors \cite{antal2019}, which can have different screen resolutions, the same computed velocity value may correspond to different physical cursor movement speeds across monitors, potentially introducing noise into the velocity encoding. Since velocity uses two separate velocity channels, this noise may affect both channels, potentially reducing the benefit of the additional directional information. Therefore, for DFL, this noise may outweigh the performance gain provided by the additional directional information, resulting in slightly worse performance than for speed encoding.

While we are able to demonstrate SOTA performance on the Balabit and TWOS datasets when compared to Chong et al. \cite{chong2020}, we also extended our evaluation to two more datasets, ChaoShen and DFL. A prior paper comparing the relative performance of the Balabit, ChaoShen, and DFL datasets reports mean AUC of 0.7186 and 0.8758 for ChaoShen and DFL respectively. Our results exceed these baselines, however, the segmentation method utilized between our method and theirs differs, making the relative improvement demonstrated for these datasets inconclusive.

\section{Conclusion}
\label{section:Conclusion}
In this work, we investigated the contribution of temporal information in image-based mouse dynamics authentication. We proposed image representations that encoded speed and velocity information while retaining spatial information. We evaluated these representations using both XYPlot and PWD representations under both location-aware and location-free settings across four publicly available mouse datasets. 

The results demonstrate that incorporating temporal information generally improves authentication performance compared with spatial-only representations. The benefit of temporal encoding was found to generalize across both the multilabel 2D-CNN and ViT architectures, indicating that the additional information provided by speed and velocity can be effectively leveraged to increase authentication performance. Preserving the absolute position of where the mouse cursor was located also improved performance over the location-free variant of each representation. PWD with velocity encoding was consistently found to be the strongest performing image representation, with the exception of DFL, where performance was similar to speed encoding.

Overall, our findings demonstrate that mouse trajectory images can benefit from incorporating temporal information of mouse movement over relying only on spatial characteristics. By combining spatial and temporal features within a unified image representation, the proposed approach allows vision models to leverage additional information that leads to an increase in authentication performance.

\bibliographystyle{IEEEtran}
\bibliography{refs}

\end{document}